%% file: ToArxiv.tex
\documentclass{article}
\usepackage{ijcai25}

\usepackage{times}
\usepackage{soul}
\usepackage{url}
\usepackage[hidelinks]{hyperref}
\usepackage[utf8]{inputenc}
\usepackage[small]{caption}
\usepackage{graphicx}
\usepackage{amsmath}
\usepackage{amsthm}
\usepackage{booktabs}
\usepackage{algorithm}
\usepackage{algpseudocode}
\usepackage{xspace}
\usepackage{tabularx}
\usepackage{pifont}
\usepackage{colortbl}
\usepackage{xcolor}
\definecolor{lightgrayv}{HTML}{F4F3F8} 
\definecolor{grayv}{HTML}{707070}
\usepackage{multirow}
\usepackage{arydshln}
\usepackage{makecell}

\newcommand{\eg}{\emph{e.g.,}\xspace}
\newcommand{\ie}{\emph{i.e.,}\xspace}

\newcommand{\baby}{M\textsc{d-pcc}\xspace}
\newcommand{\dataset}{\textit{CoMis}\xspace}

\title{Robust Misinformation Detection by Visiting Potential Commonsense Conflict}

\author{
Bing Wang$^{1,2}$ \and
Ximing Li$^{1,2}$\thanks{Corresponding authors} \and
Changchun Li$^{1,2}$ \and
Bingrui Zhao$^{1,2}$ \and
Bo Fu$^{3}$ \\
Renchu Guan$^{1,2}$ \And
Shengsheng Wang$^{1,2 *}$ \\
\affiliations
$^1$College of Computer Science and Technology, Jilin University \\
$^2$Key Laboratory of Symbolic Computation and Knowledge Engineering of the MoE, Jilin University \\
$^3$School of Computer Science and Artificial Intelligence, Liaoning Normal University \\
\emails
{\normalsize \{wangbing1416, liximing86, changchunli93\}@gmail.com,
fubo@lnnu.edu.cn,
\{guanrenchu, wss\}@jlu.edu.cn}
}

\begin{document}

\maketitle

\begin{abstract}
    \input{Sec_Abstract}

\end{abstract}

\input{Sec_Introduction}

\input{Sec_Method}

\input{Sec_Data}

\input{Sec_Experiment}

\input{Sec_Relatedwork}

\input{Sec_Conclusion}

\section*{Acknowledgement}
We acknowledge support for this project from the National Key R\&D Program of China (No.2021ZD0112501, No.2021ZD0112502), the National Natural Science Foundation of China (No.62276113), China Postdoctoral Science Foundation (No.2022M721321).

\bibliographystyle{named}
\bibliography{reference}

\input{Sec_Appendix}

\end{document}

%% file: Sec_Abstract.tex
The development of Internet technology has led to an increased prevalence of misinformation, causing severe negative effects across diverse domains.
To mitigate this challenge, Misinformation Detection (MD), aiming to detect online misinformation automatically, emerges as a rapidly growing research topic in the community. 
In this paper, we propose a novel plug-and-play augmentation method for the MD task, namely Misinformation Detection with Potential Commonsense Conflict (\baby). We take inspiration from the prior studies indicating that fake articles are more likely to involve commonsense conflict. Accordingly, we construct commonsense expressions for articles, serving to express potential commonsense conflicts inferred by the difference between extracted commonsense triplet and golden ones inferred by the well-established commonsense reasoning tool COMET. These expressions are then specified for each article as augmentation. Any specific MD methods can be then trained on those commonsense-augmented articles. Besides, we also collect a novel commonsense-oriented dataset named \dataset, whose all fake articles are caused by commonsense conflict. We integrate \baby with various existing MD backbones and compare them across 4 public benchmark datasets and \dataset. Empirical results demonstrate that \baby can consistently outperform the existing MD baselines.

%% file: Sec_Introduction.tex
\section{Introduction}

Over the past decades, many social media platforms \eg Twitter and Weibo, become the mainstream avenue to share information among human beings in daily life.
Unfortunately, these platforms eventually afford convenience for the dissemination of various misinformation such as fake news and rumors \cite{vosoughi2018spread,sander2022misinformation}. 
To reduce the negative effect of misinformation, how to detect them effectively and efficiently becomes the primary task in this endeavor. Accordingly, the emergent topic of \textbf{M}isinformation \textbf{D}etection (\textbf{MD}) has recent drawn increasing attention from the natural language process community \cite{ma2016detecting,zhang2021mining,sheng2022zoom,hu2023learn,wang2024why}.

Generally, cutting-edge MD works employ a variety of deep learning techniques to learn the potential semantic correlation between online articles and their corresponding veracity labels, \eg real and fake \cite{ma2016detecting,zhang2021mining,hu2023learn,zhang2024evolving}.
For example, most MD arts concentrate on designing various models to incorporate external features, \eg entity-based embeddings of named entities in an article and their corresponding descriptions \cite{dun2021kan,hu2021compare}, domain information for adapting MD models across multiple domains \cite{nan2022improving}, and emotional signals to enhance MD models by learning potential emotional patterns \cite{zhang2021mining}.

\begin{table}[t]
\centering
\renewcommand\arraystretch{1.25}
  \caption{Real-world misinformation examples with commonsense conflict. The text fragments implying commonsense conflict are underlined. Human beings are more likely to identify these articles contain misinformation owing to the commonsense conflicts.}
  \label{examples}
  \small
  \setlength{\tabcolsep}{5pt}{
  \begin{tabular}{m{8.0cm}}
    \toprule
    \ding{202} \textbf{Article}: The body will produce toxins at any time, and if they accumulate too much, you will get sick. \underline{Drinking more juice} will help to \underline{eliminate toxins}. \\
    \textbf{Veracity label}: \textit{Fake} \\
    \hline
    \ding{203} \textbf{Article}: \underline{Meat floss} is made of \underline{cotton}. This was discovered by my niece’s mother-in-law. Moms, please pay attention. \\
    \textbf{Veracity label}: \textit{Fake} \\
    \bottomrule
  \end{tabular} }
\end{table}

Despite the success of learning the pattern between articles and veracity labels from data, we are particularly interested in, as complicated phenomenons, \textbf{how do human beings identify misinformation?} Recent psychological and sociological studies partially offer a certain kind of answer as human beings naturally distinguish misinformation by referring to their pre-existing commonsense knowledge \cite{lewandowsky2012misinformation,scheufele2019science}. In certain scenarios, articles with misinformation are more likely to involve \textbf{commonsense conflict}, and human beings will identify misinformation by leveraging, at least referring to, such conflict involved, as examples illustrated in Table~\ref{examples}.
 
To identify misinformation by simulating the way of human thinking regarding commonsense conflict, the primary problem is how to measure and express them for given articles. Accordingly, we propose a novel plug-and-play augmentation method for the MD task, namely \textbf{M}isinformation \textbf{D}etection with \textbf{P}otential \textbf{C}ommonsense \textbf{C}onflict (\textbf{\baby}). Specifically, we propose to measure the commonsense conflicts of articles by the difference between the extracted commonsense triplet and the golden triplet inferred by the well-established commonsense reasoning tool \cite{bosselut2019comet,hwang2021comet}, and use those triplets to specify a predefined \textit{commonsense template} as \textit{commonsense expressions} to express the potential commonsense conflicts. For each article, we integrate it with its corresponding specific commonsense expression to form an augmented one, named \textit{commonsense-augmented article}. Given those augmented articles, one can build effective detectors by any existing MD methods and backbones.

For empirical evaluations, we employ 4 public benchmark datasets \textit{GossipCop} \cite{shu2020fakenewsnet}, \textit{Weibo} \cite{sheng2022zoom}, \textit{PolitiFact} \cite{shu2020fakenewsnet} and \textit{Snopes} \cite{popat2017where}. Additionally, we further collect a new \textbf{C}ommonsense-\textbf{o}riented \textbf{Mis}information benchmark datasets, named \textbf{\dataset}, whose all fake articles are caused by commonsense conflict. We integrate \baby with various existing MD backbones and compare them across public benchmark datasets and \dataset. Empirical results demonstrate that \baby can consistently outperform the existing MD baselines. Our source code and data are released in \url{https://github.com/wangbing1416/MD-PCC}.

The primary contributions of this paper can be summarized as the following three-folds:
\begin{itemize}
    \item We propose a plug-and-play augmentation MD method, named \baby, by expressing the potential commonsense conflict.
    \item We collect a new commonsense-oriented misinformation dataset, named \dataset, whose all fake articles are caused by commonsense conflict.
    \item We conduct experiments across both public benchmark datasets and \dataset, and empirical results indicate the effectiveness of \baby.
\end{itemize}

%% file: Sec_Method.tex
\section{Proposed \baby Method} \label{sec2}
In this section, we briefly review several preliminaries including the task definition of MD and prevalent commonsense reasoning methods. We then describe the proposed method \baby in more detail.

\begin{figure*}[t]
  \centering
  \includegraphics[scale=0.95]{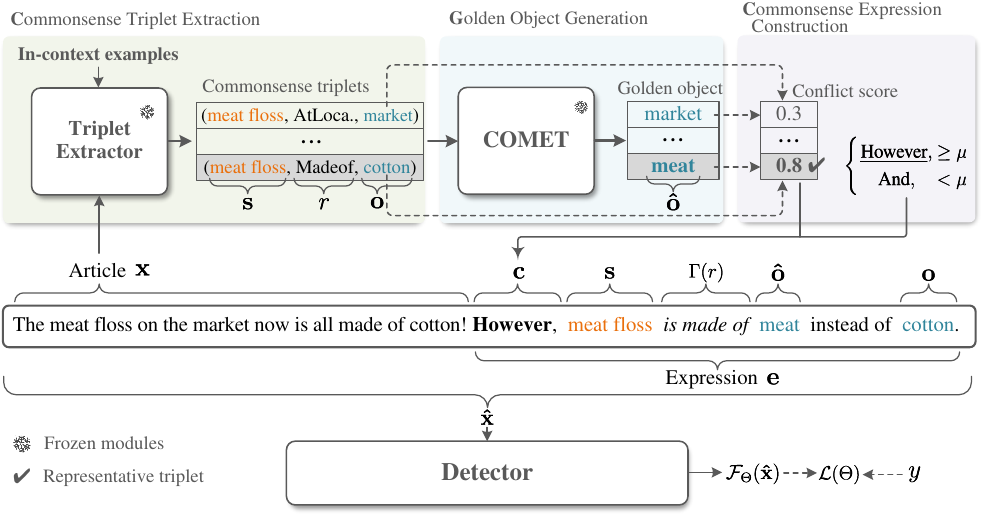}
  \caption{The overall framework of \textbf{\baby}. Its basic idea is to construct a commonsense expression $\mathbf{e}$ and specify it as an augmentation. To achieve this, given an article $\mathbf{x}$, we input it and several in-context examples into a triplet extractor to extract commonsense triplets. Then, we generate corresponding golden objects for them using the commonsense tool. Finally, we calculate commonsense conflict scores for each pair of extracted and golden objects, and select one with the highest score, \eg 0.8, to construct the commonsense expression. In the framework, the parameters of the triplet extractor and COMET are frozen, and the detector will be optimized with Eq.~\eqref{eq1}.}
  \label{framework}
\end{figure*}

\subsection{Preliminaries} \label{preliminary}

\paragraph{Task formulation of MD.} Commonly, the basic goal of MD is to induce a detector $\mathcal{F}_{\boldsymbol{\theta}}(\cdot)$ over a given training dataset $\mathcal{D}$, and use $\mathcal{F}_{\boldsymbol{\theta}}(\cdot)$ to distinguish whether any unseen article is real or fake. We formally describe the dataset of $N$ training samples as $\mathcal{D} = \{\mathbf{x}_i, y_i\}_{i=1}^N$, where each sample is composed of a raw article $\mathbf{x}_i$ and its corresponding veracity label $y_i \in \{0, 1\}$, \ie 0/1 indicating fake/real. With any specific detector $\mathcal{F}_{\boldsymbol{\theta}}(\cdot)$, it can be trained by optimizing the following objective with respect to ${\boldsymbol{\theta}}$:   
\begin{equation}
    \label{eq0}
    \mathcal{L}({\boldsymbol{\theta}}) = \frac{1}{N} \sum \nolimits _{i=1}^N \ell \left( \mathcal{F}_{\boldsymbol{\theta}}(\mathbf{x}_i), y_i \right),
\end{equation}
where $\ell (\cdot, \cdot)$ denotes the binary cross-entropy loss commonly.

\paragraph{Commonsense reasoning.}
Generally speaking, current commonsense reasoning methods aim to train a generative language model referring to the relation triplet $(\mathbf{s}, r, \mathbf{o})$, where $\mathbf{s}$ and $\mathbf{o}$ are the subject and object, respectively, and $r$ is the relation between them. Given any subject-relation pair $(\mathbf{s},r)$, a commonsense reasoning method can accurately predict the corresponding object $\mathbf{o}$. Typically, the methods are trained across the commonsense-oriented dataset ATOMIC$_{20}^{20}$ \cite{hwang2021comet}, which comprises a substantial collection of relation triplets. The typical commonsense-oriented relations of ATOMIC$_{20}^{20}$ include \{\texttt{xNeed}, \texttt{xAttr}, \texttt{xReact}, \texttt{xEffect}, \texttt{xWant}, \texttt{xIntent}, \texttt{oEffect}, \texttt{oReact}, \texttt{oWant}, \texttt{isAfter}, \texttt{HasSubEvent}, \texttt{HinderedBy}\}, representing the relations between specific events or human actions. Beyond these ones, these methods can also be generalized to a large knowledge base ConceptNet \cite{speer2017conceptnet}, so as to capture the relations between entities including \{\texttt{MadeOf}, \texttt{AtLocation}, \texttt{isA}, \texttt{Partof}, \texttt{HasA}, \texttt{UsedFor}\}. To make notation simple, we use $\mathcal{R}$ to denote the set of all those relations captured by the commonsense reasoning methods.

\subsection{Overview of \baby} \label{sec.overview}

Basically, our \baby is a plug-and-play augmentation method for the MD task. We take inspiration from the assumption that fake articles are more likely to involve commonsense conflict. Accordingly, we design a \textit{commonsense template} to express the potential commonsense conflict measured by prevalent commensense reasoning methods and specify it for each original article as the augmentation. To be specific, the commonsense template is designed as 
\begin{equation}
    \mathbf{c} \oplus \mathbf{s} \oplus \Gamma(r) \oplus \mathbf{\hat o}\ \big[\oplus \textit{``instead of''} \oplus \mathbf{o} \ \big], \nonumber
\end{equation}
where $(\mathbf{s}, r, \mathbf{o})$ indicates the \textit{representative commonsense triplet} extracted from the article, $\mathbf{\hat o}$ is the \textbf{golded} object corresponding to $(\mathbf{s}, r)$ generated by commonsense reasoning methods, and $\Gamma(r)$ denotes the original expression of $r$, \eg ``\textit{is made of}'' is the original expression of \texttt{MadeOf}. 
We suppose that an article involves a commonsense conflict if $\mathbf{o} \neq \mathbf{\hat o}$, otherwise $\mathbf{o} = \mathbf{\hat o}$. Accordingly, we define that $\mathbf{c}$ will be specified by the adversative conjunction word ``\textit{However}'' when $\mathbf{o} \neq \mathbf{\hat o}$; and by contrast, it will be specified by ``\textit{And}'', and the text segment $\textit{``instead of''} \oplus \mathbf{o}$ will be excluded.

With this commonsense template, for each article $\mathbf{x}_i$, we form its corresponding \textit{commonsense expression} $\mathbf{e}_i$ by specifying $(\mathbf{s}_i, r_i, \mathbf{o}_i)$, $\mathbf{\hat o}_i$ and $\mathbf{c}_i$ with three stages: \textbf{commonsense triplet extraction}, \textbf{golden object generation}, and \textbf{commonsense expression construction}, respectively. 
Accordingly, we concatenate $\mathbf{x}_i$ and $\mathbf{e}_i$ as a commonsense-augmented article $\mathbf{\hat x}_i$. Given all commonsense-augmented samples $\{\mathbf{\hat x}_i, y_i\}_{i=1}^N$, we can formulate the following objective with respect to any specific detector $\mathcal{F}_{\boldsymbol{\theta}}(\cdot)$:
\begin{equation}
    \label{eq1}
    \mathcal{L}({\boldsymbol{\theta}}) = \frac{1}{N} \sum \nolimits _{i=1}^N \ell \left( \mathcal{F}_{\boldsymbol{\theta}}(\mathbf{\hat x}_i ), y_i \right),
    \  \mathbf{\hat x}_i = \mathbf{x}_i \oplus \mathbf{e}_i.
\end{equation}
For clarity, the overall framework of \baby is depicted in Fig.~\ref{framework}. In the following subsections, we will describe the three stages of generating commonsense expressions.

\subsection{Commonsense Triplet Extraction} \label{sec2.2}

In this stage, for each article $\mathbf{x}_i$, we extract a certain number of relevant commonsense triplets $\{(\mathbf{s}_i^\gamma, r_i^\gamma, \mathbf{o}_i^\gamma)\}_{\gamma=1}^{|\mathcal{\overline R}_i|}$. To achieve this, we first screen all relations of $\mathcal{R}$ to extract all corresponding triplets $\{(\mathbf{s}_i^\gamma, r_i^\gamma, \mathbf{o}_i^\gamma)\}_{\gamma=1}^{|\mathcal{R}|}$ from $\mathbf{x}_i$ and then filter out the meaningless ones from them.

Specifically, we first extract $\{(\mathbf{s}_i^\gamma, r_i^\gamma, \mathbf{o}_i^\gamma)\}_{\gamma=1}^{|\mathcal{R}|}$ by prompting an existing LLM with the In-Context Learning (ICL) method \cite{brown2020language,min2022rethinking}. We design natural language queries $\{\mathcal{T}^\gamma\}_{\gamma=1}^{|\mathcal{R}|}$ for relations $\{r_i^\gamma\}_{\gamma=1}^{|\mathcal{R}|}$, \eg ``\textit{Extract entity1 and entity2 from the text where entity1 is made of entity2. Text:}'' for the relation \texttt{MadeOf}. Accordingly, the formulation of in-context examples $\{\mathcal{I}_k^\gamma\}_{k=1}^K$ for the relation $r^\gamma$ is delineated as: 
\begin{align}
    \label{eq2}
    \mathcal{I}_k^\gamma = \mathcal{T}^\gamma \oplus \mathbf{x}_k^\gamma \oplus \textit{entity1 is} \ \mathbf{s}_k^\gamma \ \textit{and entity2 is} & \ \mathbf{o}_k^\gamma, \nonumber \\
    \gamma \in \{1, 2, \cdots, |\mathcal{R}|\}, k \in \{1, 2, \cdots, K\} . & 
\end{align}

We collect $K$ labeled examples for each relation to facilitate ICL.
Then, we input both in-context examples and query $\mathcal{T}^\gamma \oplus \mathbf{x}_i$ into a triplet extractor $\mathcal{G}_\Phi(\cdot)$ specified by a pre-trained T5 model \cite{raffel2020exploring} to generate $\mathbf{s}_i^\gamma$ and $\mathbf{o}_i^\gamma$:
\begin{align}
    \label{eq3}
    \mathbf{s}_i^\gamma, \mathbf{o}_i^\gamma \gets \mathcal{G}_\Phi \left( \mathcal{I}_1^\gamma \oplus \cdots \oplus \mathcal{I}_K^\gamma \oplus \mathcal{T}^\gamma \oplus \mathbf{x}_i \right), & \nonumber \\
    \gamma \in \{1, 2, \cdots, |\mathcal{R}|\}. & 
\end{align}

Because an article does not always contain all relations of $\mathcal{R}$, we filter out the meaningless ones. We design a filtering method based on the conditional generation logits. It follows the spirit that generative models always output lower probabilities for its generated uncertain word tokens, which can be evaluated by \textit{perplexity} \cite{jurafsky2000speech,lee2021towards}. Specifically, we define the text generated by $\mathcal{G}_\Phi(\cdot)$ in Eq.~\eqref{eq3} as $\mathbf{t}_i^\gamma = \{t_{i1}^\gamma, t_{i2}^\gamma, \cdots, t_{iL}^\gamma\}$, where $L$ indicates the length of the text. And we remove the triplet $(\mathbf{s}_i^\gamma, r_i^\gamma, \mathbf{o}_i^\gamma)$, if
\begin{equation}
\label{eq4}
    \sum \nolimits _{j=1}^L \log P \left( t_{ij}^\gamma \big| t_{i<j}^\gamma; \Phi \right) > \epsilon, \nonumber
\end{equation}
where $\epsilon$ is a controllable hyper-parameter. After filtering, the set of commonsense relations for $\mathbf{x}_i$ is refined as $\overline{\mathcal{R}}_i \in \mathcal{R}$.

\subsection{Golden Object Generation}

Given $\{(\mathbf{s}_i^\gamma, r_i^\gamma)\}_{\gamma=1}^{|\overline{\mathcal{R}}_i|}$, we generate their golden objects $\{\mathbf{\hat o}_i^\gamma\}_{\gamma=1}^{|\overline{\mathcal{R}}_i|}$, which are aligned with real-world commonsense knowledge.
To be specific, we feed each $(\mathbf{s}_i^\gamma, r_i^\gamma)$ into the prevalent commonsense tool $\mathcal{G}_\Pi(\cdot)$ \cite{bosselut2019comet} to generate its golden object $\mathbf{\hat o}_i^\gamma$ as  
\begin{equation}
    \label{eq5}
    \mathbf{\hat o}_i^\gamma \gets \mathcal{G}_\Pi \left( \mathbf{s}_i^\gamma, r_i^\gamma \right), \ 
    \gamma \in \{1, 2, \cdots, |\overline{\mathcal{R}}_i|\}.
\end{equation}
We specially explain that because the prevalent commonsense reasoning tool has been pre-trained on a large-scale commonsense dataset ATOMIC$_{20}^{20}$, we treat $\mathbf{\hat o}_i^\gamma$ as the ground-truth knowledge of $(\mathbf{s}_i^\gamma, r_i^\gamma)$, \ie the corresponding golden object.

\subsection{Commonsense Expression Construction}

In this stage, we construct commonsense expression $\mathbf{e}_i$ by filling the commonsense template in Sec.~\ref{sec.overview} based on $\{(\mathbf{s}_i^\gamma, r_i^\gamma, \mathbf{o}_i^\gamma)\}_{\gamma=1}^{|\overline{\mathcal{R}}_i|}$ and $\{\mathbf{\hat o}_i^\gamma\}_{\gamma=1}^{|\overline{\mathcal{R}}_i|}$.
Specifically, we first compute conflict scores $\{c_i^\gamma\}_{\gamma=1}^{|\overline{\mathcal{R}}_i|}$ for each pair of $\mathbf{o}_i^\gamma$ and $\mathbf{\hat o}_i^\gamma$. 
We take inspiration from BARTS\textsc{core} \cite{yuan2021bartscore}, and present a new evaluation metric to compute the commonsense conflict score $c_i^\gamma$ during the process that we input $\mathbf{s}_i^\gamma$ and $r_i^\gamma$ into the commonsense reasoning tool $\mathcal{G}_\Pi(\cdot)$ with Eq.~\eqref{eq5}. The specific metric is as follows:
\begin{align}
    \label{eq6}
    c_i^\gamma = - \sum \nolimits _{j=1}^{\overline{L}} 
    \mathbf{o}_{ij}^\gamma \log
    \mathcal{P} \left( \mathbf{\hat o}_{ij}^\gamma \big| \mathbf{\hat o}_{i<j}^\gamma; \Pi \right), & \nonumber \\
    \gamma \in \{1, 2, \cdots, |\overline{\mathcal{R}}_i|\}, &
\end{align}
where $\overline{L}$ denotes the length of the generated $\mathbf{\hat o}_{ij}^\gamma$. 

Then, we select the highest conflict score $c_i$ from the set of $\{c_i^\gamma\}_{\gamma=1}^{|\overline{\mathcal{R}}_i|}$, and denote its corresponding representative commonsense triplet and golden object as $\{\mathbf{s}_i, r_i, \mathbf{o}_i\}$ and $\mathbf{\hat o}_i$, respectively. Accordingly, we fill them into the commonsense template to obtain the expression as follows:
\begin{equation}
\renewcommand{\arraystretch}{1.15}
    \label{eq7}
    \mathbf{e}_i = \left\{
    \setlength{\arraycolsep}{1.40pt}{
    \begin{array}{rlr}
        \textit{``However''} \oplus & \mathbf{s}_i \oplus \Gamma(r_i) \oplus \mathbf{\hat o}_i \oplus &  \\
        & \textit{``instead of''} \oplus \mathbf{o}_i, & c_i \geq \mu, \\ 
        \textit{``And''} \oplus & \mathbf{s}_i \oplus \Gamma(r_i) \oplus \mathbf{\hat o}_i,& c_i < \mu,
    \end{array} }
    \right.
\end{equation}
where $\Gamma(\cdot)$ is the original expression for each relation, \eg ``\textit{is made of}'' for the relation \texttt{MadeOf}. When $c_i \geq \mu$, we argue that the article $\mathbf{x}_i$ exists the commonsense conflict; otherwise, there is not. In summary, the training summary of \baby is presented in Alg.~\ref{algorithm}.

\renewcommand{\algorithmicrequire}{\textbf{Input:}}
\renewcommand{\algorithmicensure}{\textbf{Output:}}
\begin{algorithm}[t]
\small
    \caption{Training summary of \baby.}
    \label{algorithm}
    \begin{algorithmic}[1]
    \Require Training dataset $\mathcal{D} = \{\mathbf{x}_i, y_i\}_{i=1}^N$; pre-trained language model $\mathcal{G}_\Phi(\cdot)$; commonsense reasoning tool $\mathcal{G}_\Pi(\cdot)$; commonsense relations $\mathcal{R}$; query templates $\{\mathcal{T}^\gamma\}_{\gamma=1}^{|\mathcal{R}|}$.
    \Ensure detection model $\mathcal{F}_{\boldsymbol{\theta}}(\cdot)$; expressions $\{\mathbf{e}_i\}_{i=1}^N$.
    \For{$i = 1, 2, \cdots, N$}
        \State $\mathcal{C}_i \gets [\ ]$,
        \For{$r_i^\gamma$ in $\mathcal{R}$}
            \State extract $\mathbf{s}_i^\gamma$ and $\mathbf{o}_i^\gamma$ with $\mathcal{G}_\Phi(\cdot)$ in Eq.~\eqref{eq3},
            \If{Eq.~\eqref{eq4} is not satisfied}
            \State $\mathbf{\widehat o}_i^\gamma \gets \mathcal{G}_\Pi \left( \mathbf{s}_i^\gamma, r_i^\gamma \right)$,
            \State calculate $c_i^\gamma$ with Eq.~\eqref{eq6}, $\mathcal{C}_i \gets c_i^\gamma$.
            \EndIf
        \EndFor
        \State select $\{\mathbf{s}_i, r_i, \mathbf{o}_i\}$ and $\mathbf{\widehat o}_i$ with $\max(\mathcal{C}_i)$,
        \State construct $\mathbf{e}_i$ with Eq.~\eqref{eq7}.
    \EndFor
    \State train $\mathcal{F}_{\boldsymbol{\theta}}(\cdot)$ with $\mathcal{L}$ in Eq.~\eqref{eq1}.
    \end{algorithmic}
\end{algorithm}

%% file: Sec_Data.tex
\section{Datasets}

To evaluate the performance of \baby, we conduct experiments by employing four public MD datasets \textit{GossipCop} \cite{shu2020fakenewsnet}, \textit{Weibo} \cite{sheng2022zoom}, \textit{PolitiFact} \cite{shu2020fakenewsnet} and \textit{Snopes} \cite{popat2017where}. Additionally, we also collect a new Chinese MD dataset, referred to as \textbf{\dataset}, wherein all fake articles can be verified by leveraging commonsense conflict. 
We describe their details in the following section. For clarity, their statistics are shown in Table~\ref{datasetsta}.

\begin{table}[t]
\centering
\renewcommand\arraystretch{1.0}
\small
  \caption{Statistics of prevalent FND datasets and \dataset.}
  \label{datasetsta}
  \setlength{\tabcolsep}{5pt}{
  \begin{tabular}{m{1.50cm}<{\centering}m{0.70cm}<{\centering}m{0.70cm}<{\centering}m{0.50cm}<{\centering}m{0.70cm}<{\centering}m{0.50cm}<{\centering}m{0.70cm}<{\centering}}
    \toprule
    \multirow{2}{*}{Dataset} & \multicolumn{2}{c}{\# Train} & \multicolumn{2}{c}{\# Val.} & \multicolumn{2}{c}{\# Test} \\
    \cmidrule(r){2-3} \cmidrule(r){4-5} \cmidrule(r){6-7}
    & Fake & Real & Fake & Real & Fake & Real \\
    \hline
    \textit{Weibo} & 2,561 & 7,660 & 499 & 1,918 & 754 & 2,957 \\
    \textit{GossipCop} & 2,024 & 5,039 & 604 & 1,774 & 601 & 1,758 \\
    \textit{PolitiFact} & 1,224 & 1,344 & 170 & 186 & 307 & 337 \\
    \textit{Snopes} & 2,288 & 838 & 317 & 116 & 572 & 210 \\
    \textit{\dataset} & 560 & 440 & 170 & 125 & 162 & 123 \\
    \bottomrule
  \end{tabular} }
\end{table}

\begin{table*}[t]
\centering
\renewcommand\arraystretch{1.05}
  \caption{Experimental results of our \baby on four prevalent datasets \textit{Weibo}, \textit{GossipCop}, \textit{PolitiFact} and \textit{Snopes}. The results marked by $*$ indicate that they are statistically significant than the baseline methods (p-value \textless 0.05).}
  \label{result}
  \small
  \setlength{\tabcolsep}{5pt}{
  \begin{tabular}{m{4.2cm}m{1.55cm}<{\centering}m{1.55cm}<{\centering}m{1.55cm}<{\centering}m{1.55cm}<{\centering}m{1.55cm}<{\centering}m{1.55cm}<{\centering}m{1.0cm}<{\centering}}
    \toprule
    \quad \quad \quad \quad \quad Method & Macro F1 & Accuracy & Precision & Recall & F1$_\text{real}$ & F1$_\text{fake}$ & Avg.$\boldsymbol{\Delta}$ \\
    \hline
    \multicolumn{8}{c}{\textbf{Dataset: \textit{Weibo}}} \\
    EANN \cite{wang2018eann} & 76.53{\footnotesize \color{grayv} $\pm$0.52} & 84.62{\footnotesize \color{grayv} $\pm$0.30} & 76.75{\footnotesize \color{grayv} $\pm$0.63} & 76.07{\footnotesize \color{grayv} $\pm$1.14} & 90.43{\footnotesize \color{grayv} $\pm$0.25} & 62.41{\footnotesize \color{grayv} $\pm$1.12} & - \\
    \rowcolor{lightgrayv} \quad + \baby (ours) & 77.30{\footnotesize \color{grayv} $\pm$0.99}$^*$ & 85.88{\footnotesize \color{grayv} $\pm$0.50}$^*$ & 78.58{\footnotesize \color{grayv} $\pm$0.89}$^*$ & 76.29{\footnotesize \color{grayv} $\pm$0.89} & 91.25{\footnotesize \color{grayv} $\pm$0.32}$^*$ & 63.36{\footnotesize \color{grayv} $\pm$0.78}$^*$ & \textbf{+0.98} \\
    
    BERT \cite{devlin2019bert} & 75.64{\footnotesize \color{grayv} $\pm$0.41} & 84.13{\footnotesize \color{grayv} $\pm$0.67} & 75.58{\footnotesize \color{grayv} $\pm$1.09} & 75.79{\footnotesize \color{grayv} $\pm$0.74} & 90.02{\footnotesize \color{grayv} $\pm$0.52} & 61.26{\footnotesize \color{grayv} $\pm$0.59} & - \\
    \rowcolor{lightgrayv} \quad + \baby (ours) & 76.80{\footnotesize \color{grayv} $\pm$0.86}$^*$ & 84.62{\footnotesize \color{grayv} $\pm$0.92} & 76.32{\footnotesize \color{grayv} $\pm$1.41}$^*$ & 77.44{\footnotesize \color{grayv} $\pm$0.80}$^*$ & 90.26{\footnotesize \color{grayv} $\pm$0.67} & 63.35{\footnotesize \color{grayv} $\pm$1.16}$^*$ & \textbf{+1.06} \\
    
    BERT-EMO \cite{zhang2021mining}& 76.17{\footnotesize \color{grayv} $\pm$0.48} & 84.60{\footnotesize \color{grayv} $\pm$0.40} & 76.27{\footnotesize \color{grayv} $\pm$0.64} & 76.11{\footnotesize \color{grayv} $\pm$0.85} & 90.34{\footnotesize \color{grayv} $\pm$0.31} & 61.99{\footnotesize \color{grayv} $\pm$0.89} & - \\
    \rowcolor{lightgrayv} \quad + \baby (ours) & 77.03{\footnotesize \color{grayv} $\pm$1.21}$^*$ & 85.29{\footnotesize \color{grayv} $\pm$1.19}$^*$ & 77.50{\footnotesize \color{grayv} $\pm$1.00}$^*$ & 76.72{\footnotesize \color{grayv} $\pm$0.94}$^*$ & 91.53{\footnotesize \color{grayv} $\pm$0.80}$^*$ & 63.28{\footnotesize \color{grayv} $\pm$0.69}$^*$ & \textbf{+0.98} \\
    
    CED \cite{wu2023category} & 76.42{\footnotesize \color{grayv} $\pm$1.55} & 85.51{\footnotesize \color{grayv} $\pm$1.32} & 77.92{\footnotesize \color{grayv} $\pm$0.87} & 75.70{\footnotesize \color{grayv} $\pm$0.63} & 90.72{\footnotesize \color{grayv} $\pm$0.91} & 62.42{\footnotesize \color{grayv} $\pm$1.40} & - \\
    \rowcolor{lightgrayv} \quad + \baby (ours) & 78.33{\footnotesize \color{grayv} $\pm$0.20}$^*$ & 86.59{\footnotesize \color{grayv} $\pm$0.51}$^*$ & 79.98{\footnotesize \color{grayv} $\pm$1.22}$^*$ & 77.13{\footnotesize \color{grayv} $\pm$1.11}$^*$ & 91.70{\footnotesize \color{grayv} $\pm$0.42}$^*$ & 64.96{\footnotesize \color{grayv} $\pm$0.63}$^*$ & \textbf{+1.67} \\

    DM-INTER \cite{wang2024why} & 76.29{\footnotesize \color{grayv} $\pm$0.42} & 84.59{\footnotesize \color{grayv} $\pm$0.33} & 76.23{\footnotesize \color{grayv} $\pm$0.51} & 76.39{\footnotesize \color{grayv} $\pm$0.87} & 90.31{\footnotesize \color{grayv} $\pm$0.27} & 62.26{\footnotesize \color{grayv} $\pm$0.84} & - \\
    \rowcolor{lightgrayv} \quad + \baby (ours) & 77.59{\footnotesize \color{grayv} $\pm$0.23}$^*$ & 85.80{\footnotesize \color{grayv} $\pm$0.72}$^*$ & 78.43{\footnotesize \color{grayv} $\pm$0.77}$^*$ & 77.32{\footnotesize \color{grayv} $\pm$0.74}$^*$ & 91.15{\footnotesize \color{grayv} $\pm$0.58}$^*$ & 64.13{\footnotesize \color{grayv} $\pm$0.64}$^*$ & \textbf{+1.39} \\
    \hline
    \specialrule{0em}{0.5pt}{0.5pt}
    \hline
    \multicolumn{8}{c}{\textbf{Dataset: \textit{GossipCop}}} \\
    EANN \cite{wang2018eann} & 78.59{\footnotesize \color{grayv} $\pm$0.84} & 84.47{\footnotesize \color{grayv} $\pm$0.66} & 80.37{\footnotesize \color{grayv} $\pm$1.46} & 77.42{\footnotesize \color{grayv} $\pm$1.36} & 89.80{\footnotesize \color{grayv} $\pm$0.55} & 67.39{\footnotesize \color{grayv} $\pm$1.59} & - \\
    \rowcolor{lightgrayv} \quad + \baby (ours) & 79.80{\footnotesize \color{grayv} $\pm$0.47}$^*$ & 85.08{\footnotesize \color{grayv} $\pm$0.35}$^*$ & 80.82{\footnotesize \color{grayv} $\pm$0.86} & 79.02{\footnotesize \color{grayv} $\pm$1.05}$^*$ & 90.12{\footnotesize \color{grayv} $\pm$0.32} & 69.48{\footnotesize \color{grayv} $\pm$0.99}$^*$ & \textbf{+1.05} \\
    
    BERT \cite{devlin2019bert} & 78.23{\footnotesize \color{grayv} $\pm$0.45} & 83.78{\footnotesize \color{grayv} $\pm$0.80} & 79.00{\footnotesize \color{grayv} $\pm$1.45} & 77.49{\footnotesize \color{grayv} $\pm$0.57} & 89.21{\footnotesize \color{grayv} $\pm$0.69} & 67.24{\footnotesize \color{grayv} $\pm$0.45} & - \\
    \rowcolor{lightgrayv} \quad + \baby (ours) & 79.10{\footnotesize \color{grayv} $\pm$0.46}$^*$ & 84.61{\footnotesize \color{grayv} $\pm$0.56}$^*$ & 80.32{\footnotesize \color{grayv} $\pm$1.10}$^*$ & 78.24{\footnotesize \color{grayv} $\pm$0.47}$^*$ & 89.85{\footnotesize \color{grayv} $\pm$0.45}$^*$ & 68.37{\footnotesize \color{grayv} $\pm$0.60}$^*$ & \textbf{+0.92} \\
    
    BERT-EMO \cite{zhang2021mining} & 78.42{\footnotesize \color{grayv} $\pm$0.47} & 83.92{\footnotesize \color{grayv} $\pm$0.39} & 79.15{\footnotesize \color{grayv} $\pm$0.73} & 77.10{\footnotesize \color{grayv} $\pm$1.01} & 89.67{\footnotesize \color{grayv} $\pm$0.59} & 67.23{\footnotesize \color{grayv} $\pm$1.03} & - \\
    \rowcolor{lightgrayv} \quad + \baby (ours) & 79.32{\footnotesize \color{grayv} $\pm$0.27}$^*$ & 84.68{\footnotesize \color{grayv} $\pm$0.66}$^*$ & 80.28{\footnotesize \color{grayv} $\pm$1.38}$^*$ & 78.63{\footnotesize \color{grayv} $\pm$0.67}$^*$ & 90.03{\footnotesize \color{grayv} $\pm$0.36} & 68.81{\footnotesize \color{grayv} $\pm$0.31}$^*$ & \textbf{+1.04} \\
    
    CED \cite{wu2023category} & 78.33{\footnotesize \color{grayv} $\pm$0.40} & 83.77{\footnotesize \color{grayv} $\pm$0.68} & 78.85{\footnotesize \color{grayv} $\pm$1.26} & 77.94{\footnotesize \color{grayv} $\pm$0.25} & 89.17{\footnotesize \color{grayv} $\pm$0.57} & 67.49{\footnotesize \color{grayv} $\pm$0.25} & - \\
    \rowcolor{lightgrayv} \quad + \baby (ours) & 79.79{\footnotesize \color{grayv} $\pm$0.52}$^*$ & 85.52{\footnotesize \color{grayv} $\pm$0.31}$^*$ & 82.04{\footnotesize \color{grayv} $\pm$0.67}$^*$ & 78.23{\footnotesize \color{grayv} $\pm$0.84} & 90.54{\footnotesize \color{grayv} $\pm$0.22}$^*$ & 69.04{\footnotesize \color{grayv} $\pm$0.96}$^*$ & \textbf{+1.60} \\

    DM-INTER \cite{wang2024why} & 78.29{\footnotesize \color{grayv} $\pm$0.56} & 84.04{\footnotesize \color{grayv} $\pm$0.40} & 79.43{\footnotesize \color{grayv} $\pm$0.87} & 77.43{\footnotesize \color{grayv} $\pm$1.00} & 89.45{\footnotesize \color{grayv} $\pm$0.34} & 67.21{\footnotesize \color{grayv} $\pm$1.09} & - \\
    \rowcolor{lightgrayv} \quad + \baby (ours) & 79.76{\footnotesize \color{grayv} $\pm$0.42}$^*$ & 85.08{\footnotesize \color{grayv} $\pm$0.30}$^*$ & 80.85{\footnotesize \color{grayv} $\pm$0.75}$^*$ & 78.93{\footnotesize \color{grayv} $\pm$0.93}$^*$ & 90.13{\footnotesize \color{grayv} $\pm$0.28}$^*$ & 69.40{\footnotesize \color{grayv} $\pm$0.87}$^*$ & \textbf{+1.38} \\
    \hline
    \specialrule{0em}{0.5pt}{0.5pt}
    \hline
    \multicolumn{8}{c}{\textbf{Dataset: \textit{PolitiFact}}} \\
    BERT \cite{devlin2019bert} & 60.36{\footnotesize \color{grayv} $\pm$0.99} & 60.49{\footnotesize \color{grayv} $\pm$2.04} & 60.53{\footnotesize \color{grayv} $\pm$2.18} & 60.45{\footnotesize \color{grayv} $\pm$2.08} & 62.86{\footnotesize \color{grayv} $\pm$1.74} & 56.62{\footnotesize \color{grayv} $\pm$2.25} & - \\
    \rowcolor{lightgrayv} \quad + \baby (ours) & 61.92{\footnotesize \color{grayv} $\pm$0.68}$^*$ & 62.45{\footnotesize \color{grayv} $\pm$0.47}$^*$ & 62.46{\footnotesize \color{grayv} $\pm$0.39}$^*$ & 62.05{\footnotesize \color{grayv} $\pm$0.57}$^*$ & 66.29{\footnotesize \color{grayv} $\pm$0.46}$^*$ & 57.55{\footnotesize \color{grayv} $\pm$1.70}$^*$ & \textbf{+1.90} \\

    CED \cite{wu2023category} & 61.75{\footnotesize \color{grayv} $\pm$0.54} & 61.86{\footnotesize \color{grayv} $\pm$0.50} & 61.79{\footnotesize \color{grayv} $\pm$0.51} & 61.77{\footnotesize \color{grayv} $\pm$0.54} & 63.56{\footnotesize \color{grayv} $\pm$0.90} & 59.94{\footnotesize \color{grayv} $\pm$1.23} & - \\
    \rowcolor{lightgrayv} \quad + \baby (ours) & 63.60{\footnotesize \color{grayv} $\pm$0.21}$^*$ & 63.87{\footnotesize \color{grayv} $\pm$0.34}$^*$ & 63.84{\footnotesize \color{grayv} $\pm$0.37}$^*$ & 63.63{\footnotesize \color{grayv} $\pm$0.23}$^*$ & 66.59{\footnotesize \color{grayv} $\pm$1.28}$^*$ & 60.61{\footnotesize \color{grayv} $\pm$1.05}$^*$ & \textbf{+1.91} \\

    DM-INTER \cite{wang2024why} & 60.85{\footnotesize \color{grayv} $\pm$1.96} & 61.23{\footnotesize \color{grayv} $\pm$1.77} & 61.23{\footnotesize \color{grayv} $\pm$1.71} & 60.97{\footnotesize \color{grayv} $\pm$1.81} & 64.15{\footnotesize \color{grayv} $\pm$1.56} & 57.54{\footnotesize \color{grayv} $\pm$1.57} & - \\
    \rowcolor{lightgrayv} \quad + \baby (ours) & 63.13{\footnotesize \color{grayv} $\pm$1.58}$^*$ & 63.37{\footnotesize \color{grayv} $\pm$1.51}$^*$ & 63.29{\footnotesize \color{grayv} $\pm$1.51}$^*$ & 63.14{\footnotesize \color{grayv} $\pm$1.55}$^*$ & 66.08{\footnotesize \color{grayv} $\pm$1.28}$^*$ & 60.17{\footnotesize \color{grayv} $\pm$1.17}$^*$ & \textbf{+2.20} \\
    \hline
    \specialrule{0em}{0.5pt}{0.5pt}
    \hline
    \multicolumn{8}{c}{\textbf{Dataset: \textit{Snopes}}} \\
    BERT \cite{devlin2019bert} & 62.74{\footnotesize \color{grayv} $\pm$0.78} & 72.15{\footnotesize \color{grayv} $\pm$1.74} & 64.36{\footnotesize \color{grayv} $\pm$2.03} & 62.14{\footnotesize \color{grayv} $\pm$0.70} & 43.56{\footnotesize \color{grayv} $\pm$1.71} & 81.91{\footnotesize \color{grayv} $\pm$1.58} & - \\
    \rowcolor{lightgrayv} \quad + \baby (ours) & 64.69{\footnotesize \color{grayv} $\pm$1.36}$^*$ & 73.42{\footnotesize \color{grayv} $\pm$1.89}$^*$ & 65.99{\footnotesize \color{grayv} $\pm$1.50}$^*$ & 64.14{\footnotesize \color{grayv} $\pm$1.20}$^*$ & 47.19{\footnotesize \color{grayv} $\pm$1.48}$^*$ & 82.19{\footnotesize \color{grayv} $\pm$1.43}$^*$ & \textbf{+1.79} \\
    
    CED \cite{wu2023category} & 63.60{\footnotesize \color{grayv} $\pm$1.15} & 72.39{\footnotesize \color{grayv} $\pm$0.93} & 64.34{\footnotesize \color{grayv} $\pm$0.79} & 63.29{\footnotesize \color{grayv} $\pm$1.51} & 45.74{\footnotesize \color{grayv} $\pm$1.93} & 81.44{\footnotesize \color{grayv} $\pm$1.06} & - \\
    \rowcolor{lightgrayv} \quad + \baby (ours) & 66.41{\footnotesize \color{grayv} $\pm$1.32}$^*$ & 74.82{\footnotesize \color{grayv} $\pm$0.77}$^*$ & 67.46{\footnotesize \color{grayv} $\pm$1.02}$^*$ & 65.79{\footnotesize \color{grayv} $\pm$1.49}$^*$ & 49.61{\footnotesize \color{grayv} $\pm$1.58}$^*$ & 83.21{\footnotesize \color{grayv} $\pm$0.63}$^*$ & \textbf{+2.75} \\

    DM-INTER \cite{wang2024why} & 63.24{\footnotesize \color{grayv} $\pm$1.37} & 72.83{\footnotesize \color{grayv} $\pm$0.84} & 64.41{\footnotesize \color{grayv} $\pm$1.28} & 62.62{\footnotesize \color{grayv} $\pm$1.35} & 44.47{\footnotesize \color{grayv} $\pm$1.41} & 82.01{\footnotesize \color{grayv} $\pm$0.58} & - \\
    \rowcolor{lightgrayv} \quad + \baby (ours) & 65.79{\footnotesize \color{grayv} $\pm$1.34}$^*$ & 74.01{\footnotesize \color{grayv} $\pm$1.11}$^*$ & 66.51{\footnotesize \color{grayv} $\pm$1.39}$^*$ & 65.38{\footnotesize \color{grayv} $\pm$0.72}$^*$ & 49.06{\footnotesize \color{grayv} $\pm$1.86}$^*$ & 82.53{\footnotesize \color{grayv} $\pm$0.92} & \textbf{+2.28} \\
    \bottomrule
  \end{tabular} }
\end{table*}

\subsection{Prevalent MD Datasets}

We evaluate the method with the following four MD datasets:
\begin{itemize}
    \item \textbf{\textit{GossipCop}} and \textbf{\textit{PolitiFact}} are English MD datasets sourced from \textit{FakeNewsNet} \cite{shu2020fakenewsnet}. We divide \textit{GossipCop} based on \cite{zhu2022generalizing}, which includes articles posted between 2000 and 2017 for training, with the test set consisting of articles from 2018. For \textit{PolitiFact}, we adhere to its original dataset division.
    \item \textbf{\textit{Weibo}} \cite{sheng2022zoom} is sourced from a Chinese social media platform, and we split articles published from 2010 to 2017 allocated for training and those from 2018 used for testing.
    \item \textbf{\textit{Snopes}} \cite{popat2017where} is gathered from a well-known fact-checking website \textit{snopes.com}. We split the dataset according to its original paper.
\end{itemize}

\subsection{Our Collected \dataset}

We collect a new commonsense-oriented MD dataset \dataset with the effort of human annotators.
Table~\ref{datasetsta} provides the statistics of our newly constructed dataset \dataset. The dataset contains a total of 1,580 pieces of data entries, covering diverse domains. The domain most extensively represented in the dataset pertains to food safety. 

\paragraph{Data source.}
Our MD data is sourced from two distinct channels: pre-existing MD datasets and external websites. 

First, we select suitable data items from pre-existing datasets dedicated to fake news and rumor detection, \eg \textit{Weibo-16}, \textit{Weibo-20}, and \textit{Weibo-COVID19}.
Specifically, \textit{Weibo-16} \cite{ma2016detecting} comprises posts spanning from December 2010 to April 2014, and many duplications are meticulously filtered by \cite{zhang2021mining};
\textit{Weibo-20} \cite{zhang2021mining} extends the temporal scope of \textit{Weibo-16}, encompassing data from April 2014 to November 2018, and its labels are verified through NewsVerify\footnote{\url{https://www.newsverify.com/}};
\textit{Weibo-COVID19} \cite{lin2022detect} is collected during the surge of the COVID-19 pandemic, so all articles within this dataset are exclusively centered on COVID-19 topics.

To ensure completeness and timeliness, we also manually collect commonsense-oriented samples from two external websites. First, 
\textit{Food Rumor}\footnote{\url{http://www.xinhuanet.com/food/sppy/}} is a Chinese rumor-refuting platform, which serves as a repository for misinformation and its corresponding verification, with a predominant focus on topics related to food safety, health science, and similar domains.
Then, \textit{Science Facts}\footnote{\url{https://piyao.kepuchina.cn/}} is another Chinese platform that specializes in disseminating science popularization content, covering subjects, \eg food safety and biological science.

\paragraph{Annotation and post-process.}
The annotators are instructed to select and post-process the data items that can be verified using commonsense from the aforementioned data sources. For the data from pre-existing MD datasets, we preserve their veracity labels while systematically filtering any special symbols and website links from their content. For the data sourced from external websites, we collect fake claims in the rumor-refuting channels of these websites, and real claims from their science popularization channels. Meanwhile, we maintain a consistent average claim length of approximately 50, aligning with the standards set by existing datasets.

%% file: Sec_Experiment.tex
\section{Experimental Results}

\begin{table*}[t]
\centering
\renewcommand\arraystretch{1.05}
  \caption{Experimental results of our \baby on our constructed datasets \dataset. The results marked by $*$ indicate that they are statistically significant than the baseline methods (p-value \textless 0.05).}
  \label{oursresult}
  \small
  \setlength{\tabcolsep}{5pt}{
  \begin{tabular}{m{4.2cm}m{1.55cm}<{\centering}m{1.55cm}<{\centering}m{1.55cm}<{\centering}m{1.55cm}<{\centering}m{1.55cm}<{\centering}m{1.55cm}<{\centering}m{1.0cm}<{\centering}}
    \toprule
    \quad \quad \quad \quad \quad Method & Macro F1 & Accuracy & Precision & Recall & F1$_\text{real}$ & F1$_\text{fake}$ & Avg.$\boldsymbol{\Delta}$ \\
    \hline
    BERT \cite{devlin2019bert} & 88.70{\footnotesize \color{grayv} $\pm$0.53} & 89.02{\footnotesize \color{grayv} $\pm$0.56} & 88.90{\footnotesize \color{grayv} $\pm$0.69} & 88.54{\footnotesize \color{grayv} $\pm$0.42} & 88.22{\footnotesize \color{grayv} $\pm$0.67} & 90.60{\footnotesize \color{grayv} $\pm$0.55} & - \\
    \rowcolor{lightgrayv} \quad + \baby (ours) & 91.55{\footnotesize \color{grayv} $\pm$0.36}$^*$ & 91.71{\footnotesize \color{grayv} $\pm$0.35}$^*$ & 91.42{\footnotesize \color{grayv} $\pm$0.33}$^*$ & 91.78{\footnotesize \color{grayv} $\pm$0.48}$^*$ & 90.37{\footnotesize \color{grayv} $\pm$0.47}$^*$ & 92.72{\footnotesize \color{grayv} $\pm$0.34}$^*$ & \textbf{+2.60} \\

    CED \cite{wu2023category} & 89.22{\footnotesize \color{grayv} $\pm$1.09} & 89.58{\footnotesize \color{grayv} $\pm$1.00} & 89.82{\footnotesize \color{grayv} $\pm$0.88} & 88.88{\footnotesize \color{grayv} $\pm$1.27} & 87.31{\footnotesize \color{grayv} $\pm$1.45} & 91.12{\footnotesize \color{grayv} $\pm$0.77} & - \\
    \rowcolor{lightgrayv} \quad + \baby (ours) & 91.69{\footnotesize \color{grayv} $\pm$1.11}$^*$ & 91.86{\footnotesize \color{grayv} $\pm$1.12}$^*$ & 91.61{\footnotesize \color{grayv} $\pm$1.25}$^*$ & 91.87{\footnotesize \color{grayv} $\pm$0.93}$^*$ & 90.51{\footnotesize \color{grayv} $\pm$1.16}$^*$ & 92.78{\footnotesize \color{grayv} $\pm$1.07}$^*$ & \textbf{+2.40} \\

    DM-INTER \cite{wang2024why} & 89.34{\footnotesize \color{grayv} $\pm$0.74} & 89.57{\footnotesize \color{grayv} $\pm$0.71} & 89.24{\footnotesize \color{grayv} $\pm$0.68} & 89.47{\footnotesize \color{grayv} $\pm$0.84} & 87.78{\footnotesize \color{grayv} $\pm$0.93} & 90.90{\footnotesize \color{grayv} $\pm$0.56} & - \\
    \rowcolor{lightgrayv} \quad + \baby (ours) & 91.61{\footnotesize \color{grayv} $\pm$0.96}$^*$ & 91.81{\footnotesize \color{grayv} $\pm$0.97}$^*$ & 91.63{\footnotesize \color{grayv} $\pm$0.92}$^*$ & 91.62{\footnotesize \color{grayv} $\pm$0.80}$^*$ & 90.31{\footnotesize \color{grayv} $\pm$1.00}$^*$ & 92.90{\footnotesize \color{grayv} $\pm$0.92}$^*$ & \textbf{+2.26} \\
    \bottomrule
  \end{tabular} }
\end{table*}

\begin{table}[t]
\centering
\renewcommand\arraystretch{1.05}
  \caption{Ablative study of \baby on two datasets \textit{Weibo} and \textit{GossipCop}. w/o represents without, and $\mathbf{c}$ and $\mathbf{o}$ are conjunctions and extracted objects in commonsense expressions, respectively.}
  \label{ablative}
  \small
  \setlength{\tabcolsep}{5pt}{
  \begin{tabular}{m{1.80cm}m{0.66cm}<{\centering}m{0.66cm}<{\centering}m{0.66cm}<{\centering}m{0.66cm}<{\centering}m{0.66cm}<{\centering}m{0.66cm}<{\centering}}
    \toprule
    \quad Method & F1 & Acc. & Pre. & Rec. & F1$_\text{real}$ & F1$_\text{fake}$ \\
    \hline
    \multicolumn{7}{c}{\textbf{Dataset: \textit{Weibo}}} \\
    CED & 76.42 & 85.51 & 77.92 & 75.70 & 90.72 & 62.42 \\
    \rowcolor{lightgrayv} \quad + \baby & 78.33 & 86.59 & 79.98 & 77.13 & 91.70 & 64.96 \\
    \hdashline
    \quad w/o ICL & 76.43 & 84.84 & 76.87 & 76.39 & 90.49 & 62.38 \\
    \quad w/o $\mathbf{c}$ & 77.22 & 85.33 & 77.55 & 76.65 & 90.76 & 63.43 \\
    \quad w/o $\mathbf{o}$ & 77.45 & 85.56 & 77.87 & 77.69 & 91.00 & 64.14 \\
    \hline
    \specialrule{0em}{0.5pt}{0.5pt}
    \hline
    \multicolumn{7}{c}{\textbf{Dataset: \textit{GossipCop}}} \\
    CED & 78.33 & 83.77 & 78.85 & 77.94 & 89.17 & 67.49 \\
    \rowcolor{lightgrayv} \quad + \baby & 79.79 & 85.52 & 82.04 & 78.23 & 90.54 & 69.04 \\
    \hdashline
    \quad w/o ICL & 78.40 & 83.93 & 79.15 & 77.80 & 89.33 & 67.46 \\
    \quad w/o $\mathbf{c}$ & 78.90 & 84.85 & 81.01 & 77.41 & 90.10 & 67.69 \\
    \quad w/o $\mathbf{o}$ & 79.27 & 84.65 & 80.13 & 78.54 & 89.83 & 68.71 \\
    \bottomrule
  \end{tabular} }
\end{table}

In this section, we aim to empirically evaluate our proposed method \baby, and answer the following questions:
\begin{itemize}
    \item \textbf{Q1}: Can the proposed \baby consistently improve the performance of misinformation detectors?
    \item \textbf{Q2}: Is \baby sensitive to its hyper-parameters and primary components?
    \item \textbf{Q3}: Can the generated expression $\mathbf{e}$ expresses the commonsense conflict of the article?
\end{itemize}

\subsection{Experimental Settings}

\paragraph{Baselines.}
We evaluate our plug-and-play method \baby across five prevalent MD approaches, including \textbf{EANN} \cite{wang2018eann}, \textbf{BERT} \cite{devlin2019bert}, \textbf{BERT-EMO} \cite{zhang2021mining}, the SOTA MD model \textbf{CED} \cite{wu2023category}, and \textbf{DM-INTER} \cite{wang2024why}. 

\paragraph{Implementation Details.}
In our experiments, we employ pre-trained language models FlanT5$_\text{Large}$\footnote{\url{https://huggingface.co/google/flan-t5-large}.} \cite{chung2022scaling} and mT5$_\text{Large}$\footnote{\url{https://huggingface.co/google/mt5-large}.} \cite{xue2021mt5} to extract commonsense triplets for the English and Chinese MD datasets, respectively. 
To generate golden objects, we use COMET-ATOMIC$_{20}^{20}$\footnote{\url{https://github.com/allenai/comet-atomic-2020}.} \cite{hwang2021comet} for English datasets and \textit{comet-atomic-zh}\footnote{\url{https://huggingface.co/svjack/comet-atomic-zh}.} for Chinese datasets \textit{Weibo} and \textit{\dataset}.

During the training stage, we use an Adam optimizer with a learning rate of $7 \times 10^{-5}$ for the BERT model in baseline methods. For the other modules such as the linear classifier, we use a learning rate of $1 \times 10^{-4}$, and the batch size is consistently fixed to 64.
We also fix some other manual parameters empirically, such as $K$, $\epsilon$, and $\mu$ to 5, 0.8, and 0.6, respectively.
To avoid overfitting of detectors, we adopt an early stop strategy. This means that the training stage will stop when no better Macro F1 value appears for 10 epochs.

\subsection{Main Results (Q1)}

To answer Q1, Tables~\ref{result} and \ref{oursresult} report the performance outcomes of our method \baby on two benchmark datasets and our constructed dataset, respectively. To mitigate the influence of randomness, we repeat each experiment five times using five different seeds $\{1, 2, 3, 4, 5\}$. The standard deviations of the five replicates are also illustrated in Tables~\ref{result} and \ref{oursresult}.
Overall, our \baby method, which functions as a plug-in approach, can significantly and consistently improve the performance of the baseline models across all evaluation metrics. For example, on the \textit{Weibo} dataset, our \baby improves the overall F1 and fake news F1 scores by 1.91 and 2.54, respectively, compared to the current state-of-the-art MD method CED. Additionally, on \textit{GossipCop}, it achieves improvements of 1.46 and 3.19 in macro F1 and precision scores.
When we compare different MD datasets, we observe that \baby performs better on \dataset than on the other Chinese dataset \textit{Weibo} across most evaluation metrics. Specifically, when compared to the BERT baseline, \baby improves its macro F1 and precision scores by 1.16 and 0.74 on \textit{Weibo}, while it shows more significant improvements of 2.85 and 2.52 on \dataset. These results highlight the effectiveness of \baby in incorporating commonsense knowledge to enhance the detection of knowledge-rich misinformation.

\begin{table*}[t]
\centering
\renewcommand\arraystretch{1.02}
  \caption{Case study of \baby on the dataset \dataset.}
  \label{case}
  \footnotesize
  \setlength{\tabcolsep}{5pt}{
  \begin{tabular}{m{0.3cm}<{\centering}m{3.5cm}<{\centering}m{3.0cm}<{\centering}m{2.5cm}<{\centering}m{3.0cm}<{\centering}m{2.2cm}<{\centering}}
    \toprule
    \multicolumn{6}{l}{\makecell[l]{\textbf{Article}: Meat floss is made of cotton. This was discovered by my niece’s mother-in-law. Moms, please pay attention.}} \\
    \multicolumn{6}{l}{\textbf{Expression}: However, meat floss is made of meatloaf instead of cotton.} \\
    \hdashline
    & relation $r$ & subject $\mathbf{s}$ & object $\mathbf{o}$ & gold object $\mathbf{\widehat o}$ & conflict score $c$ \\
    \rowcolor{lightgrayv} \ding{202} & \texttt{MadeOf} & meat floss & cotton & meatloaf & \textbf{0.853} \\ 
    \ding{203} & \texttt{IsA} / \texttt{HasA} & meat floss & cotton & crew meat / eat meat & 0.728 / 0.835 \\
    \ding{204} & \texttt{AtLocation} & meat floss and cotton & - & - & - \\
    \hline
    \specialrule{0em}{0.5pt}{0.5pt}
    \hline
    \specialrule{0em}{0.5pt}{0.5pt}
    
    \multicolumn{6}{l}{\makecell[l]{\textbf{Article}: Everyone has been recommended ``anti-blue light glasses'' when they go shopping for glasses. Whether they are buying \\ 
    \quad for themselves, these glasses seem to have become a must-have.Wearing it is good for your eyes and can even prevent myopia.}} \\
    \multicolumn{6}{l}{\textbf{Expression}: However, anti-blue light glasses show the effect on getting rid of blue light instead of preventing myopia. } \\
    \hdashline
    & relation $r$ & subject $\mathbf{s}$ & object $\mathbf{o}$ & gold object $\mathbf{\widehat o}$ & conflict score $c$ \\
    \ding{202} & \texttt{isA} & anti-blue light glasses & glasses & protective eyeglasses & 0.313 \\ 
    \rowcolor{lightgrayv} \ding{203} & \texttt{xEffect} & anti-blue light glasses & prevent myopia & get rid of blue light & \textbf{0.665} \\
    \ding{204} & \texttt{HinderedBy} & PersonX has anti-blue light glasses & - & - & - \\
    \bottomrule
  \end{tabular} }
\end{table*}

\subsection{Ablative Study (Q2)} \label{sec4.3}
To investigate Q2, we implement ablative experiments to assess the effectiveness of key components in \baby. Specifically, we conduct experiments on \textit{Weibo} and \textit{GossipCop}, and present three ablative versions of CED + \baby as follows:
\begin{itemize}
    \item \textbf{\baby w/o ICL}: the version without ICL ($K = 0$) in the commonsense triplet extraction stage;
    \item \textbf{\baby w/o} $\mathbf{c}$: the version without conjunction words $\mathbf{c}$, \eg ``\textit{However}'', in commonsense expressions;
    \item \textbf{\baby w/o} $\mathbf{o}$: the version without ``\textit{instead of}'' $\oplus\ \mathbf{o}$ in commonsense expressions.
\end{itemize}

The ablative results are presented in Table~\ref{ablative}. Generally, each ablative version exhibits a decreasing trend compared to \baby, illustrating the contribution of each component in our model. The overall performance ranking of these ablative versions is w/o $\mathbf{o}$ \textgreater w/o $\mathbf{c}$ \textgreater w/o ICL. This ordering indicates that: (1) the direct impact of commonsense triplet extraction on the model's performance is significant, and in-context learning consistently enhances the extraction; (2) conjunction words $\mathbf{c}$ is more important than $\mathbf{o}$ in commonsense expressions. This is because misinformation detectors can effectively learn the pattern between conjunctions and veracity labels, \eg the pattern between ``\textit{However}'' and \textit{Fake}.

\subsection{Case Study (Q3)}

The goal of \baby is to construct commonsense expressions that express the potential commonsense conflict. Therefore, we provide some representative cases in Table~\ref{case} to evaluate the generated expressions. Specifically, we select two cases from \dataset and translate them into English versions. We observe that (1) \baby extracts commonsense triples accurately, even from relatively complex articles, \eg the second case; (2) \baby can assign a higher conflict score to the triplet that does exist the commonsense conflict; (3) our presented filtering method in Sec.~\ref{sec2.2} can indeed filter out commonsense relations that do not exist in the article, \eg \texttt{AtLocation} in the first case.

%% file: Sec_Relatedwork.tex
\section{Related Works}

In this section, we briefly review the related literature about misinformation detection and commonsense reasoning.

\subsection{Misinformation Detection}

Misinformation, \eg fake news and rumors, has had a detrimental impact on society \cite{vosoughi2018spread,zhang2023sinophobia}. As a result, it has become increasingly important to identify and detect misinformation, which is referred to as misinformation detection.
Specifically, most cutting-edge MD techniques focus on detecting misinformation based on its textual and multimodal content \cite{ying2023bootstrapping,wang2024escaping}, using advanced deep learning models \cite{ma2016detecting,shu2020fakenewsnet,wang2024harmfully,xiao2024msynfd}. These models often incorporate external features like knowledge bases \cite{dun2021kan}, emotional signals \cite{zhang2021mining,jiang2024what}, and user feedback \cite{ma2016detecting,lin2023zero}. Meanwhile, some recent works have also explored strategies to leverage pre-trained large models for MD \cite{hu2023bad,chen2024can,nan2024let,wan2024dell}.

In this study, we integrate commonsense knowledge into MD models. Prior to our work, certain research efforts have aimed to leverage knowledge graphs for the enhancement of MD models. These endeavors have primarily involved learning knowledge embeddings \cite{dun2021kan,sun2022ddgcn} or retrieving entity descriptions \cite{hu2021compare,jiang2022fake}. In contrast to these approaches, our incorporation of commonsense knowledge aligns more closely with human reasoning and reactions. Meanwhile, we employ generative models for data augmentation explicitly, which obviates the need for extensive retrieval from large knowledge bases and reduces computational complexity.

\subsection{Commonsense Bases and Reasoning}

Commonsense knowledge bases, such as ConceptNet \cite{speer2017conceptnet}, ATOMIC$_{20}^{20}$ \cite{hwang2021comet}, offer a valuable resource for direct reasoning with commonsense knowledge and have found applications in various academic topics, \eg machine translation \cite{liu2023revisiting}, question answering \cite{wang2023elaboration,chen2023distinguish} and sarcasm detection \cite{min2023just}.
Recently, especially within the context of Large Language Models (LLMs), the utilization of commonsense reasoning with LLMs has garnered significant attention \cite{liu2023revisiting,shen2024the,wang2024candle}. 
These works frequently treat commonsense knowledge as supplementary information or assess the presence of commonsense knowledge within LLMs.

%% file: Sec_Conclusion.tex
\section{Conclusion}

In this paper, we aim to enhance MD models by uncovering commonsense conflicts. To achieve this goal, we propose a novel MD method named \baby, designed to generate commonsense expressions for each article, explicitly expressing commonsense conflict existing inherent in articles, and leverage it to augment original articles. 
Specifically, the expression is constructed through a commonsense triplet extracted from the original article, the corresponding golden object, and a conjunction word. To obtain these components, we first prompt the pre-trained language model with in-context examples to extract triplets and filter out irrelevant triplets. Then, the commonsense tool is employed to generate their corresponding golden objects. Finally, a new metric is designed to measure the commonsense conflict, and the conjunction word is determined using this metric. Additionally, we also collect a new commonsense-oriented MD dataset, and extensive experimental results on the datasets are conducted and prove the effectiveness of our proposed \baby.

%% file: Sec_Appendix.tex
\appendix

\section{More Details of \dataset} \label{appendixdataset}

This section offers additional details regarding our curated commonsense-oriented MD dataset \textit{\dataset}. The sources contributing to the dataset and the corresponding sample quantities are presented in Table~\ref{source}. Notably, both external websites, Science Facts and Food Rumor, serve as platforms dedicated to refuting rumors, resulting in a notable surplus of fake claims over real claims. Nevertheless, the dataset still maintains a consistent ratio of 1.3:1 between fake and real samples.

\begin{table}[h]
\centering
\renewcommand\arraystretch{1.15}
  \caption{The sources of our benchmark dataset \dataset.}
  \label{source}
  \small
  \setlength{\tabcolsep}{5pt}{
  \begin{tabular}{m{4.4cm}<{\centering}m{0.75cm}<{\centering}m{0.55cm}<{\centering}m{0.55cm}<{\centering}}
    \toprule
    Source & \#Num. & fake & real \\
    \midrule
    \textit{Weibo-16} \cite{ma2016detecting} & 523 & 223 & 300 \\
    \textit{Weibo-20} \cite{zhang2021mining} & 567 & 312 & 255 \\
    \textit{Weibo-COVID19} \cite{lin2022detect} & 69 & 22 & 47 \\
    Science Facts & 313 & 258 & 55 \\
    Food Rumor & 108 & 77 & 31 \\
    \hdashline
    \rowcolor{lightgrayv} Total & 1,580 & 892 & 688 \\
    \bottomrule
  \end{tabular} }
\end{table}

Table~\ref{datasetcase} demonstrates some cases from \textit{\dataset}, wherein we have translated Chinese content into corresponding English versions. Verification of all these cases can rely on commonsense knowledge. For instance, Example~\ding{204} can be distilled into a commonsense triplet \{PersonX beats mosquito corpses, \texttt{xEffect}, PersonX has a fungal infection\}, which can be confirmed using the COMET model.

\begin{table}[ht]
\centering
\renewcommand\arraystretch{1.15}
  \caption{Some representative examples in our dataset \textit{\dataset}.}
  \label{datasetcase}
  \small
  \setlength{\tabcolsep}{5pt}{
  \begin{tabular}{m{8.0cm}}
    \toprule
    \ding{202} \textbf{article}: The parasites in sashimi are not terrible, as long as you dip them in wasabi before eating, they can be eliminated. \\
    \rowcolor{lightgrayv} \textbf{label}: fake \quad \textbf{source}: science facts \\
    \hline
    \ding{203} \textbf{article}: Lotus root starch is a powder made from lotus root. It has a unique taste and nutritional value and is a very popular food. Although lotus root starch is good, it should be consumed in moderation to avoid excessive intake.\\
    \rowcolor{lightgrayv} \textbf{label}: real \quad \textbf{source}: science facts \\
    \hline
    \ding{204} \textbf{article}: The New England Journal of Medicine reminds: The remains of beaten mosquito corpses may enter the skin, causing fungal infections and even death! \\
    \rowcolor{lightgrayv} \textbf{label}: fake \quad \textbf{source}: \textit{Weibo-16} \\
    \hline
    \ding{205} \textbf{article}: [Learn Important Knowledge [Microphone] Do you really understand what the new coronavirus is?] There is generally no new coronavirus in the air, it is mainly spread by respiratory droplets at close range! \\
    \rowcolor{lightgrayv} \textbf{label}: real \quad \textbf{source}: \textit{Weibo-COVID19} \\
    \bottomrule
  \end{tabular} }
\end{table}

\section{Query Designing} \label{prompt}

To effectively guide large language models, \eg T5, for generating accurate article commonsense triplets, we manually design natural language queries $\{\mathcal{T}^\gamma\}_{\gamma=1}^R$ for each commonsense relation. Specifically, the prompts for event-level relations drawn from ATOMIC$_{20}^{20}$ follow the approach presented by \cite{sarik2023accent}, and are shown in Table~\ref{promptevent}. For the entity-level relations drawn from ConceptNet, their corresponding prompts are manually collected and presented in Table~\ref{promptentity}.

\begin{table}[t]
\centering
\renewcommand\arraystretch{1.15}
  \caption{Designed prompts for relations in ATOMIC$_{20}^{20}$.}
  \label{promptevent}
  \small
  \setlength{\tabcolsep}{5pt}{
  \begin{tabular}{m{1.8cm}<{\centering}m{6.0cm}}
    \toprule
    relation & prompt \\
    \midrule
     & Extract event1 and event2 from the text where {\color{gray} \texttt{[prompt]}}. Text: \\
     \hdashline
    \texttt{xNeed} & event2 needs to be true for event1 to take place \\
    \rowcolor{lightgrayv} \texttt{xAttr} & event2 shows how PersonX is viewed as after event1 \\
    \texttt{xReact} & event2 shows how PersonX reacts to event1 \\
    \rowcolor{lightgrayv} \texttt{xEffect} & event2 shows the effect of event1 on PersonX \\
    \texttt{xWant} & event2 shows what PersonX wants after event1 happens \\
    \rowcolor{lightgrayv} \texttt{xIntent} & event2 shows PersonX's intent for event1 \\
    \texttt{oEffect} & event2 shows the effect of event1 on PersonY \\
    \rowcolor{lightgrayv} \texttt{oReact} & event2 shows how PersonY reacts to event1 \\
    \texttt{oWant} & event2 shows what PersonY wants after event1 happens \\
    \rowcolor{lightgrayv} \texttt{isAfter} & event1 happens after event2 \\
    \texttt{HasSubEvent} & event1 includes event2 \\
    \rowcolor{lightgrayv} \texttt{HinderedBy} & event1 fails to happen because event2 \\
    \bottomrule
  \end{tabular} }
\end{table}

\begin{table}[h]
\centering
\renewcommand\arraystretch{1.15}
  \caption{Designed prompts for relations in ConceptNet.}
  \label{promptentity}
  \small
  \setlength{\tabcolsep}{5pt}{
  \begin{tabular}{m{1.8cm}<{\centering}m{6.0cm}}
    \toprule
    relation & prompt \\
    \midrule
     & Extract entity1 and entity2 from the text where {\color{gray} \texttt{[prompt]}}. Text: \\
     \hdashline
    \texttt{MadeOf} & entity1 is made of entity2 \\
    \rowcolor{lightgrayv} \texttt{AtLocation} & entity1 is located at entity2 \\
    \texttt{isA} & entity1 is entity2 \\
    \rowcolor{lightgrayv} \texttt{Partof} & entity1 is a part of entity2 \\
    \texttt{HasA} & entity1 has entity2 \\
    \rowcolor{lightgrayv} \texttt{UsedFor} & entity1 is used for entity2 \\
    \bottomrule
  \end{tabular} }
\end{table}

\begin{figure*}[t]
  \centering
  \includegraphics[scale=0.80]{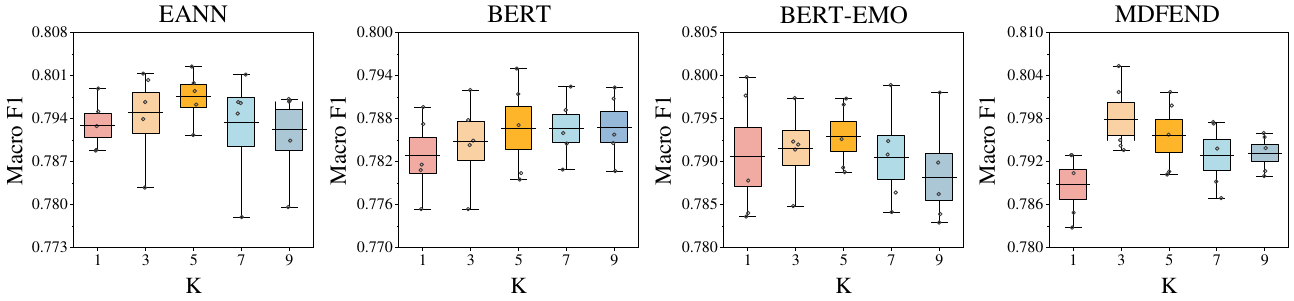}
  \caption{Sensitivity analysis of the number of in-context examples $K$.}
  \label{sensitivity}
\end{figure*}

\begin{table}[h!]
\centering
\renewcommand\arraystretch{1.15}
  \caption{In-context examples for relations in ConceptNet.}
  \label{incontextexample}
  \small
  \setlength{\tabcolsep}{5pt}{
  \begin{tabular}{m{8.0cm}}
    \toprule
    \textbf{relation}: \texttt{MadeOf} \\
    \textbf{prompt}: Extract entity1 and entity2 from the text where entity1 is made of entity2. Text: \\
    \textbf{in-context examples}: \\
    \ding{202} Meat floss is made of cotton. This was discovered by my niece’s mother-in-law. Moms, please pay attention. \textit{entity1 is meat floss and entity2 is cotton.} \\
    \ding{203} It’s amazing. I went to the supermarket and bought a package of seaweed, but it turned out to be made of plastic. It’s terrible. \textit{entity1 is seaweed and entity2 is plastic.} \\
    \ding{204} A united states is made up of 50 states. \textit{entity1 is united states and entity2 is 50 states.} \\
    \ding{205} The saxophone is considered a woodwind, even though it can be made of brass. \textit{entity1 is saxophone and entity2 is brass.} \\
    \hline
    \textbf{relation}: \texttt{AtLocation} \\
    \textbf{prompt}: Extract entity1 and entity2 from the text where entity1 is located at entity2. Text: \\
    \textbf{in-context examples}: \\
    \ding{202} You are likely to find a movie ticket around in your pants pocket weeks later. \textit{entity1 is movie ticket and entity2 is your pants pocket.} \\
    \ding{203} You are likely to find the United States in located above Mexico and below Canada. \textit{entity1 is United States and entity2 is above Mexico and below Canada.} \\
    \ding{204} Mailboxes are located at many places on streets to allow people to send mail conveniently by walking to the mailbox. \textit{entity1 is mailboxes and entity2 is many places on streets.} \\
    \ding{205} A professor can teach at a university. \textit{entity1 is professor and entity2 is university.} \\
    \bottomrule
  \end{tabular} }
\end{table}

\section{In-context Example Collection} \label{incontext}

To enhance the efficacy of extracting article commonsense triplets, we implement an in-context learning technique and curate in-context examples manually. For event-level relations in ATOMIC$_{20}^{20}$, we generate in-context examples utilizing the few-shot training samples; For entity-level relations in ConceptNet, we retrieve raw sentences from ConceptNet\footnote{\url{https://github.com/commonsense/conceptnet5/wiki/Downloads}} and the MD datasets \textit{GossipCop} and \textit{Weibo}, and choose 20 sentences for each relation to serve as their respective in-context examples. A subset of these examples is illustrated in Table~\ref{incontextexample}.

\section{More Experimental Results} \label{moreexperiments}
In this section, we will introduce more implementation details of compared baseline models and more sensitivity analysis results.

\subsection{Baselines}
We introduce the implementation details of our baselines as follows:

\begin{itemize}
    \item \textbf{EANN} \cite{wang2018eann} represents a classical multimodal MD method, employing an adversarial network to derive event-invariant features. In this work, we eliminate the image-based modality and substitute the original static GloVe embeddings with the pre-trained BERT model.
    \item \textbf{BERT} \cite{devlin2019bert} is a widely-adopted pre-trained language model. To avoid over-fitting, we freeze the parameters of its initial ten Transformer layers and solely tune its 11th layer.
    \item \textbf{BERT-EMO} \cite{zhang2021mining} stands as a misinformation detector, which incorporates emotional signals from news publishers and readers. Since the dataset used lacks comments from readers, we remove the emotional features of the reader in this work.
    \item \textbf{CED} \cite{wu2023category} emerges as the current state-of-the-art MD method. It employs an encoder for contextual representation extraction and a decoder for generating category-differentiated features. Due to the unavailability of the released code, we endeavor to replicate it on the datasets we use. 
    \item \textbf{DM-INTER} \cite{wang2024why} is a data augmentation method that introduces intent features of article publishers by reasoning on an intent hierarchy. We implement it using an LLM to reason the intent sequences and concatenate their semantic features to the article's semantic features.
\end{itemize}

\subsection{Computational Budget}
We implement our experiments on one NVIDIA GeForce 4090 GPU with 24G memory. It takes approximately 5 minutes to train a misinformation detector $\mathcal{F}_{\boldsymbol{\theta}}(\cdot)$, and 6 hours per 10,000 articles to generate commonsense expressions. Since the parameters of our used T5 model and COMET are frozen, the generating process only consumes approximately 7G memory.

\subsection{Sensitivity Analysis (Q2)} \label{appendixsensitivity}

As Sec.4.3, the number of in-context examples $K$ is an important parameter to decide the quality of extracting commonsense triplets. To this end, we conduct a sensitivity analysis on $K \in$ \{1, 3, 5, 7, 9\}, and its empirical results are depicted in Fig.~\ref{sensitivity}. The experiments are implemented across all four MD baseline models using the \textit{GossipCop} dataset. 
Our observations indicate that \baby achieves its optimal F1 score when $K$ is set to 3 or 5. Furthermore, the results consistently reveal a downward trend when $K$ either expands or shrinks. Even under some settings, the model's performance falls below the performance of the baseline model without \baby. This signifies that sub-optimal extraction of commonsense triplets can introduce additional noise, which affects the overall model performance.